\documentclass[letterpaper, 10 pt, conference]{ieeeconf}  

\IEEEoverridecommandlockouts                              

\usepackage{cite}
\usepackage{graphicx} 
\usepackage[hidelinks]{hyperref}
\usepackage{subcaption}
\usepackage{float}

\title{\LARGE \bf
A Lightweight Toggleable Adhesion Prototype for Multirotor UAV Landing on Tilting Platforms
}

\author{Teighin Nordholt and Melissa Greeff%
\thanks{The authors are with Robora Lab (www.roboralab.com), Queen’s University; and affiliated with Ingenuity Labs Research Institute. E-mails: teighin.nordholt@queensu.ca, 
melissa.greeff@queensu.ca.}
}

\begin{document}

\maketitle{}
\thispagestyle{empty}
\pagestyle{empty}

\begin{abstract}
Autonomous multirotor landings on uncrewed surface vessels (USVs) are critical for persistent maritime operations but remain challenging due to wave-induced tilt, wind disturbances, and limited landing area. Many existing approaches exhibit small pose tolerance for reliable landing. This paper presents a lightweight toggleable adhesion mechanism to improve landing reliability. The system uses a motor-driven corkscrew that engages hook-and-loop material on the landing surface, enabling active adhesion during landing and controlled release during takeoff. We evaluate a prototype using a modified Crazyflie 2.0 and a custom tilting platform at fixed angles representative of extreme wave conditions. Using only a simple vertical PID controller, the proposed approach increases landing success from an average of 40\% (baseline) to 80\% across platform tilts up to 43$^\circ$ using appropriately selected actuation settings.

\end{abstract}

\section{INTRODUCTION}

Reliable and autonomous landings of multirotor unmanned aerial vehicles (UAV) on uncrewed surface vessels (USV) in harsh conditions remain a persistent challenge. Environmental disturbances such as wave-induced tilt and wind, combined with limited pose tolerance at touchdown, can render control-intensive solutions brittle in practice \cite{PANJAVARNAM}. While significant research has focused on improving sensing, localization, and model-based control, less attention has been given to lightweight mechanisms that increase capture robustness under imprecise contact.

This work explores the development of a simple toggleable adhesion mechanism that modifies both the UAV and landing surface in a lightweight manner to improve landing reliability. Rather than relying on complex perception or control strategies, the proposed mechanism provides actively toggleable adhesion between the UAV and platform through a motor-driven corkscrew engaging with hook-and-loop material. The goal is not to replace advanced control, but to complement it by enlarging the capture envelope.

\begin{figure}
    \centering
    \includegraphics[width=0.5\textwidth, trim=0 110 0 330, 
    clip]{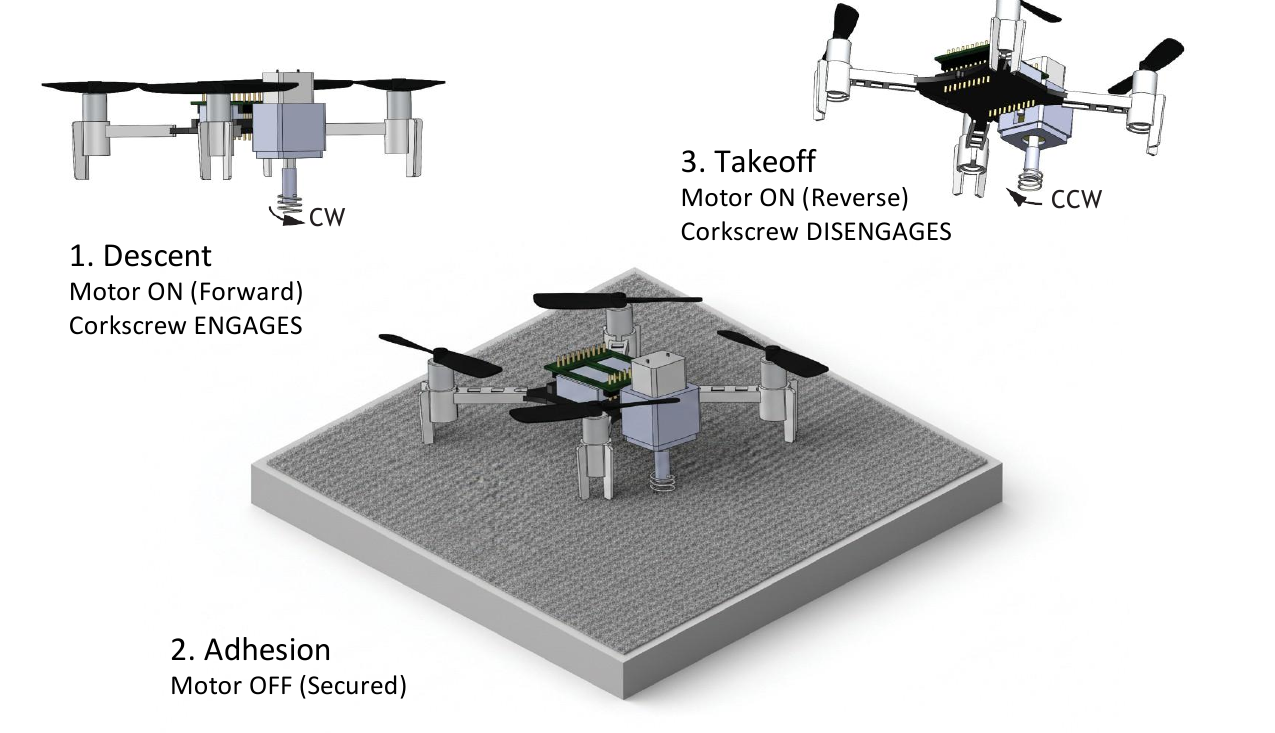}
    \caption{Operational sequence of the proposed toggleable adhesion mechanism. 
    (1) \textbf{Descent:} the onboard actuator rotates in the forward (CW) direction, engaging the corkscrew-actuated adhesion interface prior to touchdown. 
    (2) \textbf{Adhesion:} upon surface contact, the actuator is powered off and the mechanism remains mechanically secured through passive corkscrew retention. 
    (3) \textbf{Takeoff:} the actuator reverses (CCW), unloading the corkscrew and disengaging the adhesion interface to enable controlled release and lift-off.}
    \label{fig:toggleable_adhesion_sequence}
\end{figure}

To isolate mechanical performance, we intentionally adopt a reduced-assumption experimental setup. A Bitcraze Crazyflie 2.0 UAV, see Fig. \ref{fig:toggleable_adhesion_sequence}, lands on a platform fixed at varying tilt angles representative of extreme surface orientations. Landing and takeoff are executed using only a simple vertical PID controller, without wind or tilt compensation. This prototype study evaluates the feasibility, robustness, and repeatability of the mechanism under imprecise contact conditions, providing an initial exploration of mechanically assisted landing for tilted platforms. 

\section{RELATED WORK}

Reliable UAV landing in harsh conditions has been approached through two complementary research directions: control-based methods and mechanism-based solutions.

\subsection{Controls-Based Approaches}

Control-based UAV landing on dynamic platforms relies on accurate relative localization, estimation, and tracking between two independently moving agents under strong disturbances. In marine settings, wind gusts perturb the UAV while wave-driven heave and attitude changes move the USV in ways that are difficult to predict online. Sensor noise and occlusions, communication latency, and limited onboard computation further degrade closed-loop reliability. Unlike ground landings, the touchdown region is small and time-varying, leaving little tolerance for pose error and making robust UAV–USV landing a challenging control problem.

Various control strategies have been proposed, most commonly utilizing vision-based localization of the landing target \cite{FALANGA, XIA2022105288, KEIPOUR, MORANDO, GUPTA, NOVAK}. Many approaches simplify the problem by neglecting the platform's attitude, executing the landing via trajectory planning or tracking against a primarily translational target \cite{FALANGA, XIA2022105288, KEIPOUR, MORANDO, NOVAK}. Methods that explicitly account for platform tilt have largely adopted model predictive control (MPC) formulations \cite{GUPTA, BEREZA}, reflecting the importance of wave-driven attitude variations for USV deck landings. Recent work, including our own, shows that MPC-based strategies can substantially improve landing performance on dynamically tilting platforms by incorporating key aspects of surface-vessel motion into the prediction model \cite{STEPHENSON}. Nevertheless, these methods remain sensitive to sensing and modeling errors, can be computationally demanding, and do not inherently reduce the risk of vehicle damage when touchdown is imperfect. This motivates complementary mechanical solutions that enlarge the capture envelope and improve tolerance to imprecision, reducing reliance on control performance alone.

\subsection{Mechanism-Based Approaches}

Mechanism-based approaches can be grouped into three categories: charging-focused docking, capture/stabilization mechanisms, and helicopter-style deck securing systems. 

Charging-focused docking mechanisms primarily target post-landing autonomy—recharging, storage, and repeatable docking—rather than improving touchdown itself. These systems typically assume a successful landing and then use features such as post-landing repositioning on a flat pad \cite{NIU, FOX, GODZDANKER, MINNICK}, wireless or pad-based charging interfaces \cite{HOANG, SAVIOLO, CAMPI, ALJASMI, NIU, GE}, or assisted electrical connection concepts (e.g., tether/electromagnet) to increase charging robustness \cite{STUHNE}. As a result, their core performance metric is charging reliability after contact, not tolerance to imprecise touchdown on a disturbed deck.

A second class explicitly targets touchdown through mechanical capture and stabilization. Representative examples include guided mating interfaces using positive/negative coupling features \cite{PAPAYANOPOULOS, STUHNE}, active capture via manipulators or grabbing mechanisms \cite{NARVAEZ, FEDOSEEV}, stabilized or tilting landing surfaces \cite{SI, ALLURI}, and tether-based capture systems \cite{SAVIOLO, ALARCON}. These approaches show that mechanical assistance can enlarge the capture envelope and reduce failures due to timing and pose error. However, many systems are validated under controlled conditions, require substantial modification to the UAV and/or landing infrastructure, or depend on active platform hardware that is difficult to deploy on small surface vessels.

Several of these systems demonstrate improvements in specific metrics but leave key gaps for small UAV–USV applications. For example, docking interfaces and tether-based approaches can achieve centimeter-level touchdown accuracy or constrained contact thresholds, often requiring precise alignment prior to engagement \cite{ALARCON, PAPAYANOPOULOS}. Manipulator- or gripper-based systems can generate significant holding forces (e.g., hundreds of newtons in electromagnet-assisted designs \cite{NARVAEZ}) but at the cost of added mass, actuated platform components, or complex perception and coordination. Stabilized landing platforms actively compensate for tilt (often targeting deck angles on the order of tens of degrees \cite{ALLURI}), yet introduce additional degrees of actuation and mechanical complexity on the vessel side. Consequently, most mechanism-based landing systems emphasize \textit{repeatable} docking and therefore report metrics tied to approach precision-e.g., centimeter-level touchdown accuracy, tracking error, or platform-leveling performance-often under controlled conditions where the landing surface remains near-level. In contrast, our evaluation targets a different regime: capture \textit{robustness} under poor touchdown conditions.

Finally, helicopter-style shipboard securing systems provide important historical context for mechanically assisted landings on moving decks. Since the 1960s, navies have fielded mature solutions such as HARPOON (a deck-embedded capture mechanism that engages on touchdown) and BEARTRAP (a cable-and-winch mechanism that secures the helicopter during hover/approach), demonstrating that rapid mechanical restraint can be decisive in rough seas \cite{BATTELLE}. While effective, these systems assume substantial ship infrastructure and often involve manual procedures, and modern work in this area still emphasizes sensing and control to improve approach and touchdown \cite{TOPCZEWSKI, YANG}. Their success motivates UAV-scale mechanisms, but direct transfer is limited by scale, payload, and integration constraints. This reinforces the need for lightweight, mechanically simple capture concepts tailored to UAV–USV operations \cite{SAVIOLO, ALARCON}.

Prior mechanism-based systems either require substantial platform infrastructure (e.g., actuated decks/manipulators) or still depend on precise touchdown geometry, and few works quantify landing reliability as a function of platform tilt. Our work targets a lightweight UAV and surface modification that explicitly increases capture success under large fixed tilt using only simple control.

\subsection{Toggleable Adhesion}

Toggleable adhesion enables secure attachment with deliberate release, and has been explored in robotics to support temporary anchoring, climbing, and grasping. In aerial robotics, an analogous capability could improve landing robustness by providing a high-friction attachment at touchdown--tolerating pose error and disturbances--while still allowing controlled disengagement to enable reliable takeoff.

Much of the prior literature achieves toggleable adhesion through engineered materials whose adhesion is switched using stimuli such as voltage, heat, or strain \cite{RUFFATTO, YIN, JOHANNES}. While promising, these approaches are often costly, difficult to manufacture or integrate, and may provide limited load capacity for aerial landing scenarios \cite{RUFFATTO, YIN, JOHANNES}. Related ``universal" gripping concepts, such as granular jamming grippers \cite{BROWN}, can be similarly challenging to implement on small UAVs due to strength-to-weight limitations and added mechanical complexity \cite{GOMEZ}.

Mechanically simpler alternatives demonstrate that strong, toggleable attachment can be achieved via engagement with hook-and-loop surfaces. In wall-climbing robotics, for example, one approach developed an actuated end-effector terminated by metal corkscrews that engage hook-and-loop material under motor rotation and disengage via reverse rotation \cite{MALLEY}. Similar engagement/disengagement behavior has been realized using electrically actuated mechanisms such as shape memory alloys \cite{AFRISAL}. Variants of hook-based engagement have also been applied to aerial vehicles for perching, for example, one method used retractable hooks to enable fixed-wing UAVs to perch on textured surfaces \cite{DESBIENS}. Perching systems are designed for deliberate attachment to a known surface using a tightly coordinated approach maneuver, where contact geometry and timing are carefully controlled. In contrast, our problem focuses on tolerant capture during touchdown on a tilted surface under pose error and disturbance, where the mechanism must compensate for imprecise contact and forces rather than relying on precise maneuver execution.

Overall, toggleable adhesion has been studied primarily in materials science and specialized climbing/perching contexts. Its use as a lightweight landing interface explicitly intended to enlarge the capture envelope and improve robustness to imprecise contact on disturbed platforms remains relatively underexplored, motivating the investigation pursued in this work.

\section{METHODOLOGY}

\subsection{Corkscrew Gripper Design}

The corkscrew engagement concept in \cite{MALLEY} was adapted to a small multirotor landing mechanism. A helical metal “corkscrew” is rigidly coupled to the shaft of a lightweight DC gearmotor and mounted beneath the UAV such that it contacts the landing surface first. When rotated, the corkscrew progressively interlocks with hook-and-loop material on the platform, engaging multiple hooks along its length and distributing load over a larger contact region. Detachment requires commanded reverse rotation, which provides passive resistance to gravity-driven sliding on tilted surfaces while still enabling deterministic release for takeoff.

\paragraph{Corkscrew geometry selection} Before integrating the actuator, we evaluated corkscrew geometries to balance holding strength and engagement reliability. Corkscrews were fabricated from 0.5mm stainless-steel spring wire wound on mandrels of varying diameter/pitch. Each sample was engaged into hook-and-loop material using a consistent insertion procedure and then pulled quasi-statically in the surface-normal direction until disengagement. Peak pull-off force and qualitative failure modes were recorded to guide design selection. Smaller diameters generally increased peak pull-off force due to reduced elastic deformation, but frequently produced more aggressive failure (shearing the hook-and-loop interface). Larger diameters were slightly weaker in peak pull-off, but engaged more consistently by contacting and capturing a greater number of hooks. Based on these trade-offs, we selected a corkscrew of approximately 6.5mm diameter with three turns for the final prototype.

\paragraph{Actuation and integration} The corkscrew was mounted to a low-speed DC gearmotor using a custom 3D-printed fixture. A geared motor was selected to provide higher torque at low mass, reducing the likelihood of stall during engagement. Motor direction and speed were controlled using a standard H-bridge driven by two PWM signals.

\begin{figure*}[t]
    \centering
    \begin{subfigure}[t]{0.48\textwidth}
        \centering
        \includegraphics[width=\linewidth]{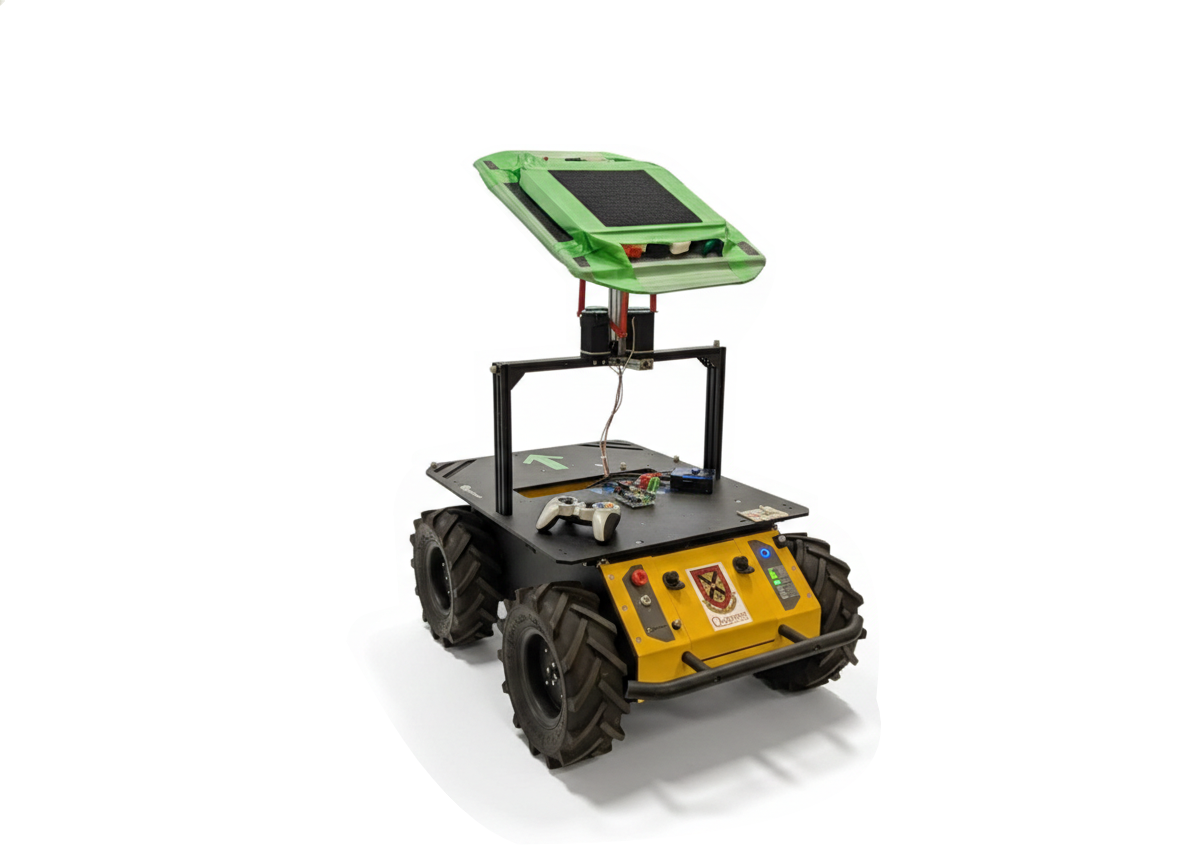}
        \caption{Experimental tilting platform with hook-and-loop surface.}
        \label{fig:exp_setup}
    \end{subfigure}
    \hfill
    \begin{subfigure}[t]{0.48\textwidth}
        \centering
        \includegraphics[width=\linewidth]{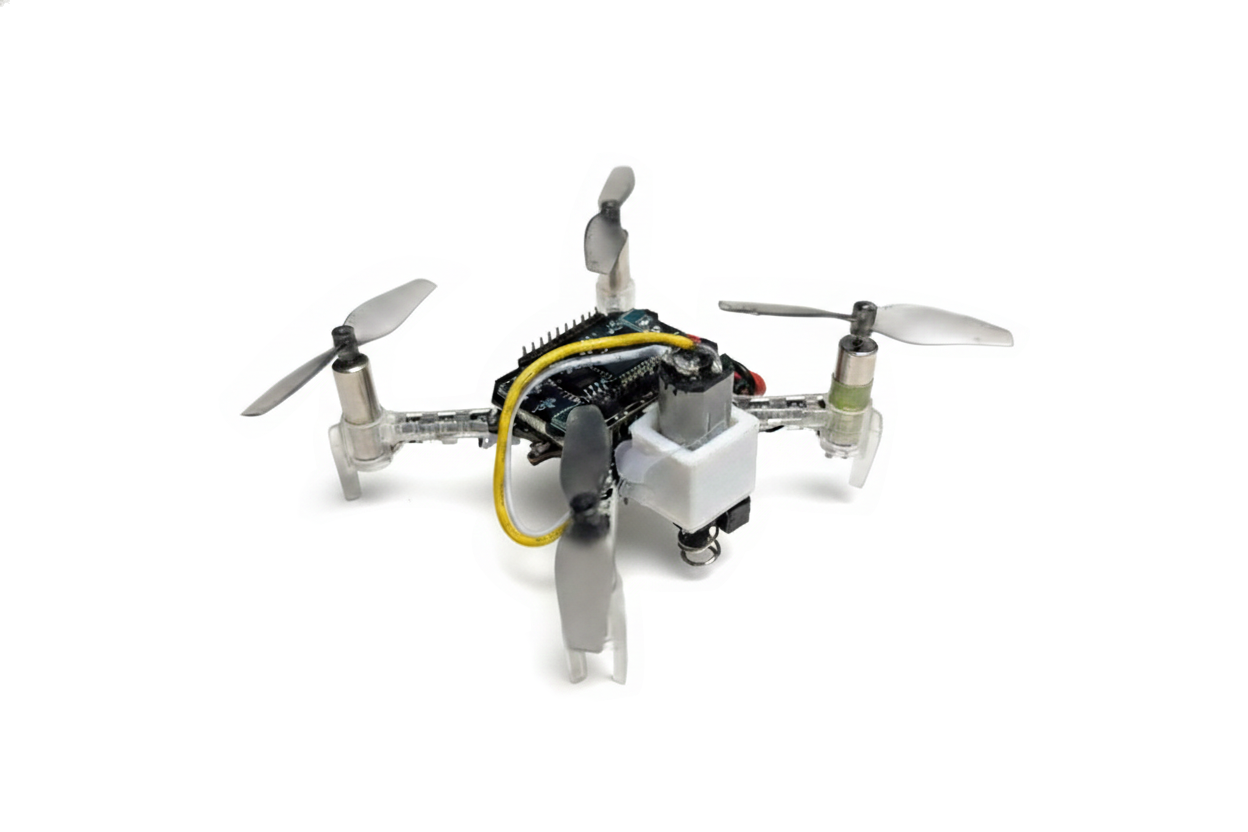}
        \caption{Crazyflie UAV with toggleable corkscrew adhesion mechanism.}
        \label{fig:mechanism_closeup}
    \end{subfigure}
    \caption{Experimental setup used to evaluate landing and takeoff robustness under fixed platform tilt.}
    \label{fig:two_column_setup}
\end{figure*}

The mechanism was integrated on a Bitcraze Crazyflie~2.0 due to its open-source firmware and rapid prototyping ecosystem. A mass budget of 20 g was allocated to limit changes in flight dynamics; the assembled mechanism added just over 10 g. The Crazyflie firmware was modified to output two PWM signals to its GPIO to command the H-bridge, and the motor assembly was mounted with a second 3D-printed bracket to ensure the corkscrew makes first contact during landing. All additional electronics were powered directly from the onboard 1S 250 mAh LiPo. The modified UAV is shown in Fig.~\ref{fig:mechanism_closeup}.

\begin{table}[t]
\centering
\caption{Test conditions for evaluating landing and takeoff performance.}
\label{tab:test_params}
\renewcommand{\arraystretch}{1.05}
\begin{tabular*}{\columnwidth}{@{\extracolsep{\fill}}l l@{}}
\hline
\textbf{Parameter} & \textbf{Values tested} \\
\hline
Platform tilt ($^\circ$) & 12, 22, 33, 43 \\
Descent speed (m/s) & $-0.25$, $-0.50$ \\
Motor duty cycle (\%) & 15, 45, 75, 100 \\
\hline
\end{tabular*}
\end{table}

\subsection{Tilting Platform Experimental Evaluation}

To evaluate the adhesion mechanism in isolation, we used the custom tilting platform from \cite{STEPHENSON} and covered its surface with hook-and-loop material (Fig.~\ref{fig:exp_setup}). The platform was commanded to fixed tilt angles (in $^\circ$ relative to horizontal) for each trial. The platform was not localized and the UAV received no information about platform pose. 
Landing and takeoff performance was characterized across all combinations of platform tilt, commanded descent speed, and corkscrew motor duty cycle listed in Table~\ref{tab:test_params}. The 100\% duty-cycle condition was tested only at 12$^\circ$ with a descent speed of $-0.25$~m/s.

\paragraph{Landing protocol} Each landing trial followed the sequence:
\begin{enumerate}
\item The UAV initializes on the ground adjacent to the platform.
\item The UAV takes off and ascends to a fixed height above the platform.
\item The UAV translates to a nominal platform-center setpoint while holding altitude.
\item The corkscrew motor is commanded to rotate at the prescribed duty cycle in the engagement direction.
\item The UAV descends at a constant commanded vertical speed for a fixed duration.
\item All motors are shut off.
\end{enumerate}
A landing is considered successful if (i) the UAV does not crash and (ii) the corkscrew is mechanically engaged with the hook-and-loop surface such that removal requires reversing the motor direction.

\paragraph{Takeoff protocol} For takeoff trials, the UAV was manually pre-engaged with the hook-and-loop surface by two turns to provide a consistent initial attachment comparable to post-landing engagement. The UAV then executed the reverse sequence (motor reversal to disengage followed by takeoff). A takeoff is considered successful if (i) the UAV does not crash and (ii) the corkscrew fully disengages, allowing the UAV to lift off from the platform.

For each parameter combination, five trials were conducted to enable broad coverage within time constraints. Baseline trials followed the same landing/takeoff procedure for each platform tilt and descent/ascent speed using an unmodified Crazyflie (no mechanism attached).

\section{EXPERIMENTAL RESULTS}

\subsection{Landing}
\begin{figure}[t]
\centering

\begin{subfigure}[b]{\columnwidth}
    \centering
    \includegraphics[width=\columnwidth]{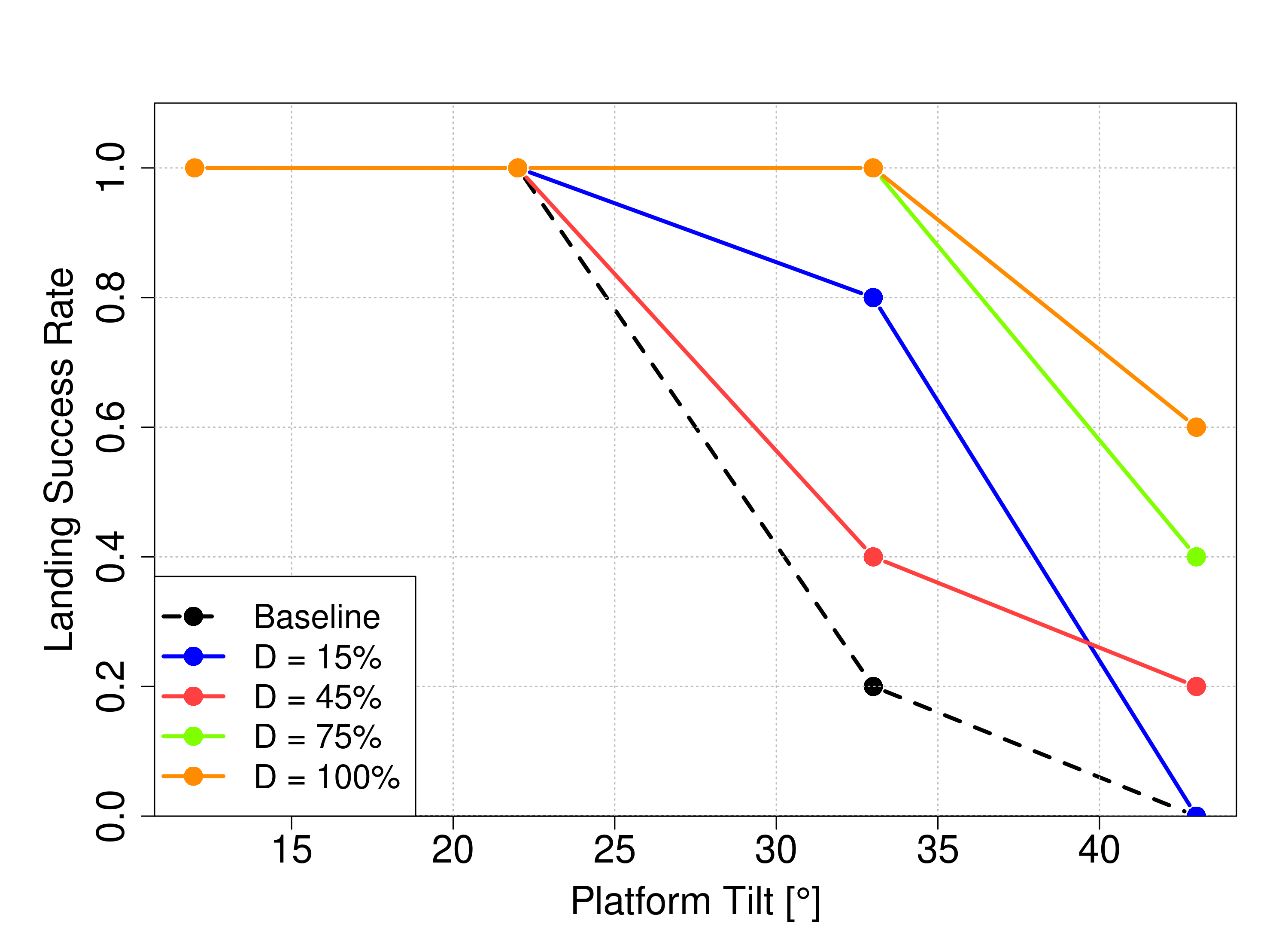}
    \caption{Descent speed = $-0.25$ m/s}
    \label{fig:landing_a}
\end{subfigure}

\begin{subfigure}[b]{\columnwidth}
    \centering
    \includegraphics[width=\columnwidth]{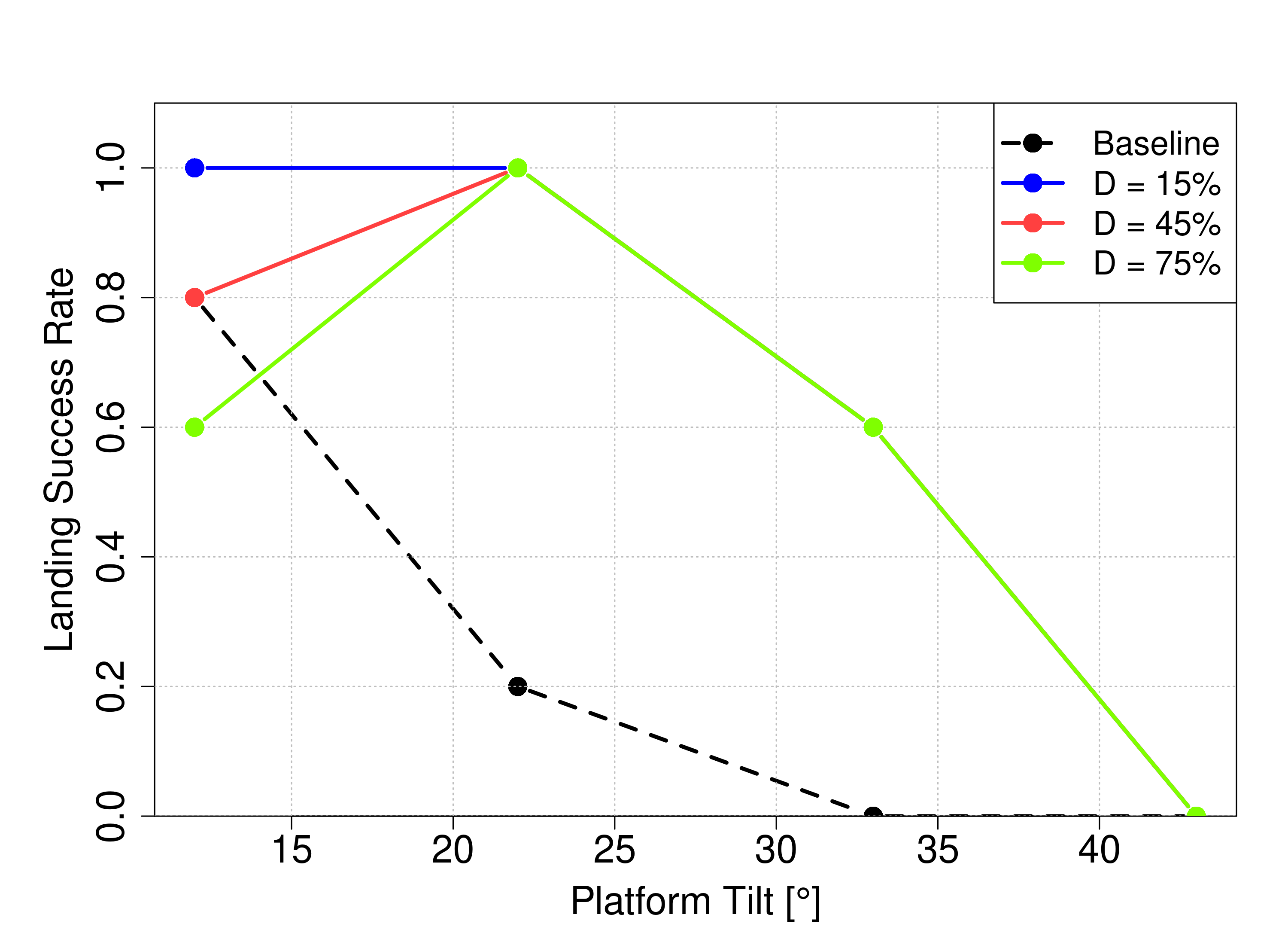}
    \caption{Descent speed = $-0.5$ m/s}
    \label{fig:landing_b}
\end{subfigure}

\caption{Landing success rate (five trials per condition) as a function of platform tilt and corkscrew motor duty cycle ($D$) at a descent speed of -0.25 m/s (a) and -0.5 m/s (b). Success is defined as the UAV not crashing with corkscrew engagement requiring reverse motor actuation to disengage. The baseline case represents an unmodified UAV.}
\label{fig:landing_combined}

\end{figure}

Landing success as a function of platform tilt, descent speed, and corkscrew motor duty cycle is shown in Fig.~\ref{fig:landing_combined}. Results are compared against a baseline case with no adhesion mechanism attached.

At a descent speed of -0.25 m/s, the baseline case exhibits rapidly decreasing landing success as platform tilt increases, dropping from perfect success at 12$^\circ$ to zero success at 43$^\circ$. In contrast, the corkscrew mechanism significantly improves landing robustness across all tested duty cycles. At moderate tilts, all duty cycles tested achieve a 100\% landing success rate. At higher tilts, increased motor duty cycle generally leads to improved performance. Even at the most extreme tilt of 43$^\circ$, successful landing is achieved for higher duty cycles, with a maximum success rate of 60\% observed at 100\% duty cycle.

At a higher descent speed of -0.5 m/s, landing success decreases for all configurations, including the baseline case. The baseline case again shows a sharp decline in performance, achieving zero successful landings at platform tilts of 33$^\circ$ and above. The corkscrew mechanism continues to outperform the baseline, particularly at intermediate tilts, though the corkscrew’s duty cycle seems to have little impact at larger tilts.

Across both descent speeds, the adhesion mechanism significantly increases landing success relative to an unmodified UAV, particularly at moderate to high platform tilts and with higher descent speeds. Higher corkscrew motor duty cycles generally improve engagement reliability, especially under more challenging landing conditions involving increased tilt. However, at a higher descent speed, the duty cycle of the corkscrew had little impact on the landing success. Therefore it seems critical that at lower descent speeds, the corkscrew should be driven quickly, but if landing faster, the corkscrew speed is not as influential to landing success.

\subsection{Takeoff}

\begin{figure}[t]
\centering

\begin{subfigure}[b]{\columnwidth}
    \centering
    \includegraphics[width=\columnwidth]{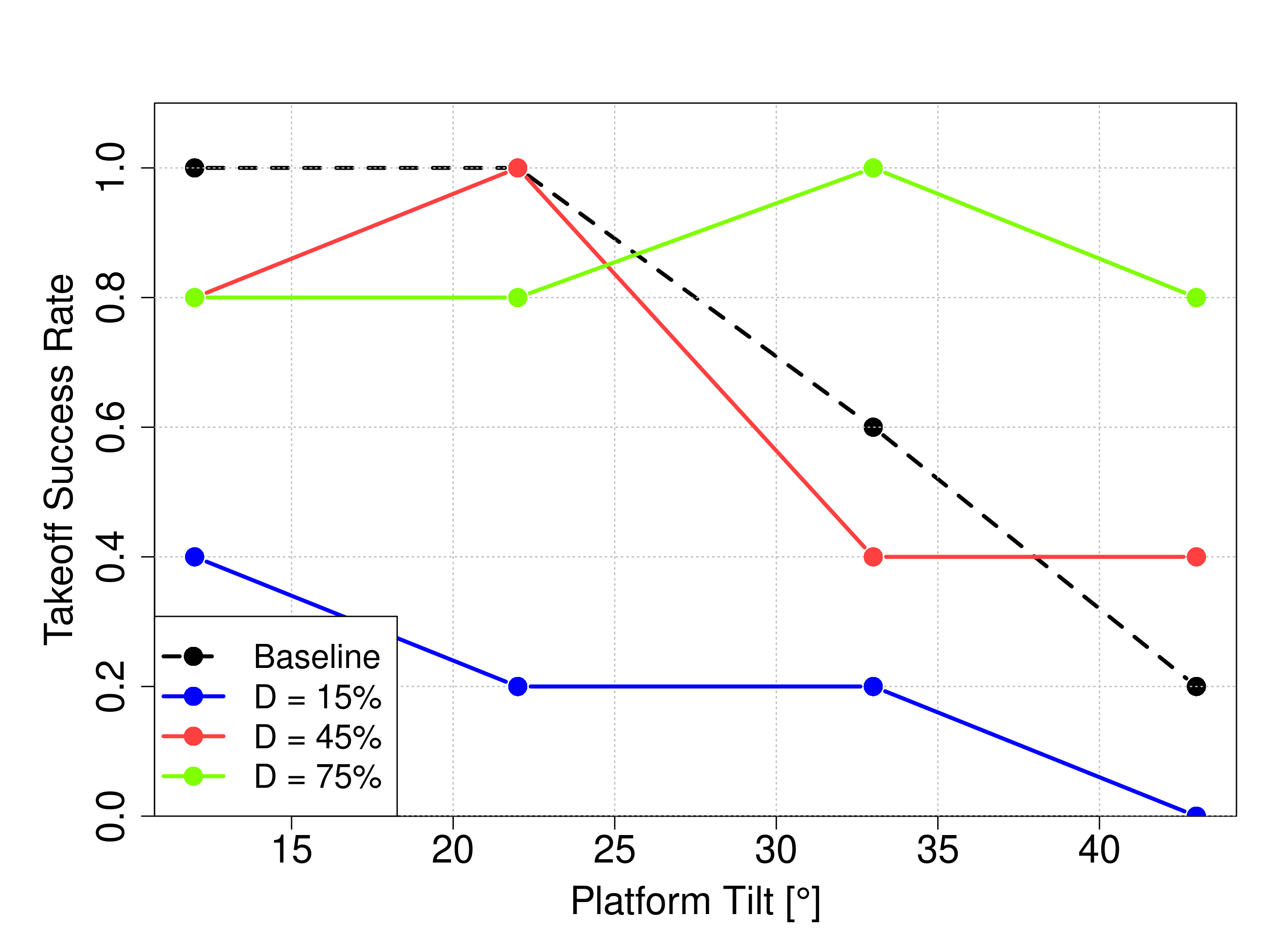}
    \caption{Takeoff speed = $0.25$ m/s}
    \label{fig:takeoff_a}
\end{subfigure}

\begin{subfigure}[b]{\columnwidth}
    \centering
    \includegraphics[width=\columnwidth]{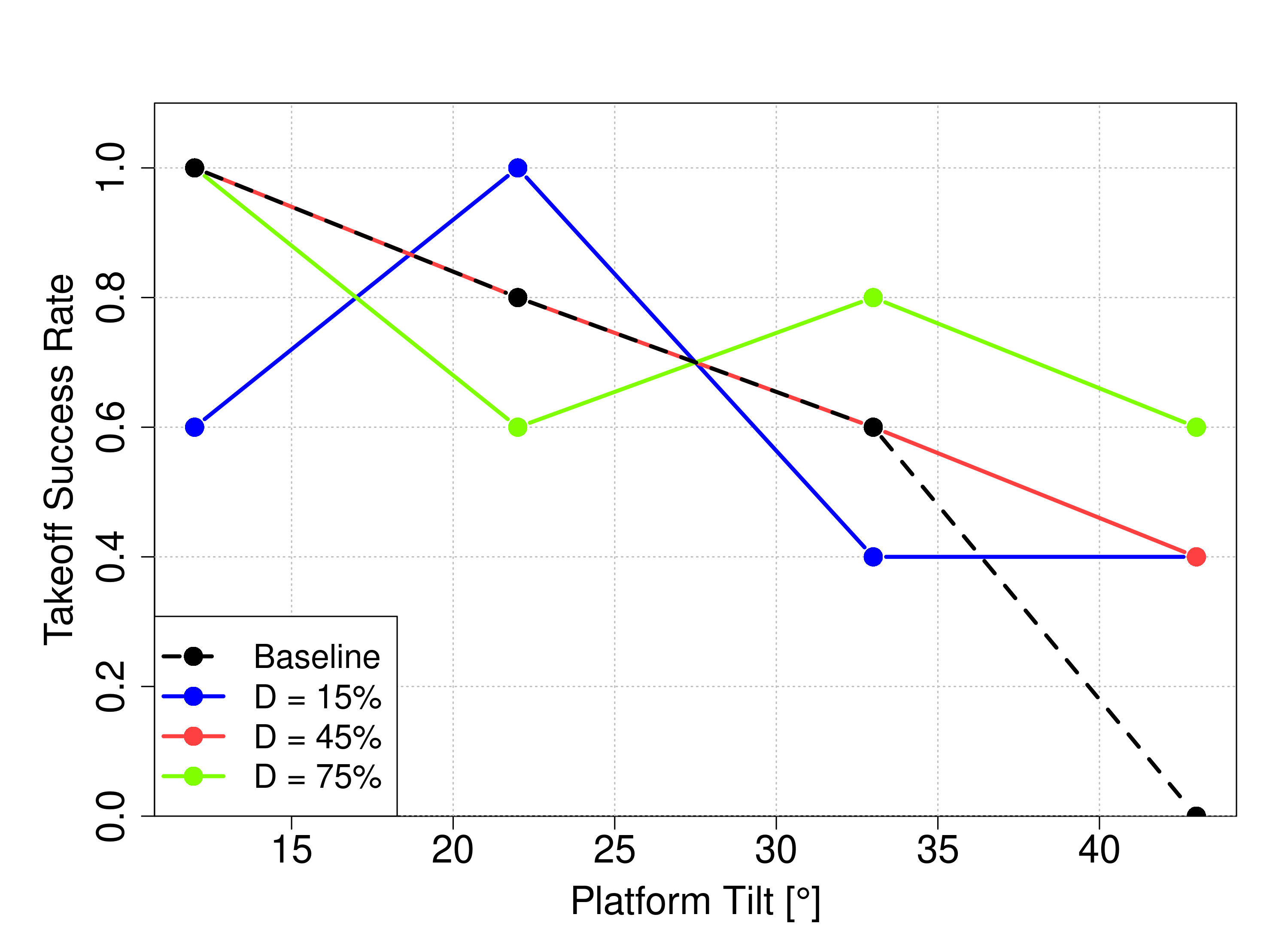}
    \caption{Takeoff speed = $-0.5$ m/s}
    \label{fig:takeoff_b}
\end{subfigure}

\caption{Takeoff success rate (five trials per condition) as a function of platform tilt and corkscrew motor duty cycle ($D$) at a takeoff speed of 0.25 m/s (a) and 0.5 m/s (b). Success is defined as the UAV not crashing and the corkscrew disengaging such that the UAV can take off. The baseline case represents an unmodified UAV. The legend in (b) applies to (a).}
\label{fig:takeoff_combined}

\end{figure}

Fig.~\ref{fig:takeoff_combined} shows the takeoff success rates characterized at both descent speeds across the platform tilt and corkscrew control speed.

Unlike with the landings, takeoffs at 0.25 m/s fail significantly more  with the mechanism compared to the baseline, specifically at low platform tilt and low motor control speed. At a tilt of 12$^\circ$, the baseline case achieves a 100\% success rate, while the lowest corkscrew duty cycle achieves only 40\% success. This was a result of the corkscrew staying engaged with the hook-and-loop material for too long, failing to disengage quick enough resulting in the UAV becoming unstable and crashing. This effect was seen less as the corkscrew was driven faster, as it was able to quickly disengage and take off.

The takeoff results at 0.5 m/s support this; the fastest corkscrew rotation outperforms the others, including the baseline, at all platform tilts but one. Notably though, the takeoff success rate with lower corkscrew speeds relative to the baseline case is improved. A trend with takeoffs is more clear, in that takeoff success rate is increased with having a higher corkscrew rotation speed or UAV takeoff speed. 

\section{CONCLUSION AND FUTURE WORK}
\addtolength{\textheight}{-0.5cm} 

This work presents an initial prototype of a lightweight, mechanically simple toggleable adhesion mechanism to improve UAV landings on tilted platforms without relying on advanced sensing or control. The proposed corkscrew-based interface engages hook-and-loop material during descent and actively disengages during takeoff, providing controlled attachment on inclined surfaces. Experiments on a fixed-tilt platform show substantially improved landing robustness over an unmodified baseline: at $-0.25$ m/s, baseline success drops from 100\% at 12$^\circ$ to 0\% at 43$^\circ$, while the mechanism achieves 100\% success at moderate tilts and up to 60\% at 43$^\circ$ with higher duty cycles; at $-0.5$ m/s, the baseline falls to 0\% by 33$^\circ$, whereas the mechanism maintains improved performance across intermediate tilts. Takeoff results further show that reliable release depends on sufficiently fast reverse actuation—low duty cycles can delay disengagement and reduce success—highlighting the need to co-design engagement and release dynamics. Overall, these results indicate that mechanically assisted adhesion can enlarge the capture envelope on highly tilted surfaces under simple PID control, motivating follow-on validation under time-varying platform motion and larger-scale vehicles.

Future work will explore two primary directions. First, scaling strategies will be investigated to enable deployment on larger UAVs typical of maritime missions. Potential design extensions include incorporating multiple corkscrews per gripper as in \cite{MALLEY} to increase effective adhesion force by engaging a larger number of loops; distributing multiple grippers across the UAV to improve load sharing, provide redundancy, and reduce sensitivity to touchdown orientation; and increasing corkscrew diameter and length to enlarge contact area and engagement depth. These modifications introduce trade-offs in stiffness, engagement dynamics, mass, and motor torque requirements, necessitating co-optimization of geometry, actuation speed, and structural integration for a given vehicle mass and expected tilt envelope. Second, future work will investigate co-design of the mechanism and flight control. While this study intentionally isolates mechanical performance under minimal PID control assumptions, integrating the adhesion mechanism with advanced control strategies—such as tilt-aware descent planning, contact-aware MPC, or adaptive engagement timing—may further enlarge the capture envelope and improve takeoff reliability. 

\bibliographystyle{IEEEtran}
\bibliography{sources.bib}

\end{document}